\pdfoutput=1
\documentclass[lettersize,journal]{IEEEtran}
\usepackage{amsmath,amsfonts,amssymb}
\usepackage{array}
\usepackage{textcomp}
\usepackage{stfloats}
\usepackage{url}
\usepackage{graphicx}
\usepackage{cite}
\usepackage{booktabs}
\usepackage{multirow}
\usepackage{orcidlink}
\usepackage{xcolor}
\usepackage{tikz}
\hypersetup{hidelinks}
\newcommand{\rsNomLo}{11.3}
\newcommand{\rsNomHi}{31.1}
\newcommand{\rsDualGCLo}{0.50}
\newcommand{\rsDualGCHi}{1.11}
\newcommand{\rsDualGCGtwoLo}{0.79}
\newcommand{\rsDualGCGtwoHi}{1.11}
\newcommand{\rsDualGCGmaxLo}{0.50}
\newcommand{\rsDualGCGmaxHi}{0.72}
\newcommand{\rsFreeGtwoG}{80}
\newcommand{\rsFreeGtwoGC}{104}
\newcommand{\rsFreeGmaxG}{64}
\newcommand{\rsFreeGmaxGC}{80}
\newcommand{\rsGammaLo}{70}
\newcommand{\rsGammaHi}{94}
\newcommand{\rsSeqN}{16}
\newcommand{\rsSeqWorseN}{15}
\newcommand{\rsSeqSigN}{12}
\newcommand{\rsSeqHi}{0.88}
\newcommand{\rsSeqGCSigN}{7}
\newcommand{\rsSeqGCN}{8}
\newcommand{\rsIndN}{16}
\newcommand{\rsIndBetterN}{1}
\newcommand{\rsIndWorseN}{1}
\newcommand{\rsIndMaxDiff}{0.20}
\newcommand{\rsIndGCLoadedL}{$-$0.13 [$-$0.18, $-$0.07]}
\newcommand{\rsIndGCLoadedR}{+0.09 [+0.02, +0.16]}
\newcommand{\rsIndPosGtwoLo}{0.83}
\newcommand{\rsIndPosGtwoHi}{1.61}
\newcommand{\rsIndRotGtwoLo}{1.55}
\newcommand{\rsIndRotGtwoHi}{1.71}
\newcommand{\rsIndPosGmaxLo}{1.83}
\newcommand{\rsIndPosGmaxHi}{3.03}
\newcommand{\rsIndRotGmaxLo}{0.70}
\newcommand{\rsIndRotGmaxHi}{1.00}
\newcommand{\rsIndOneWaistLo}{1.03}
\newcommand{\rsIndOneWaistHi}{3.21}
\newcommand{\rsIndOwnWaistLo}{0.45}
\newcommand{\rsIndOwnWaistHi}{1.05}
\newcommand{\rsLcN}{25}
\newcommand{\rsLcNmin}{20}
\newcommand{\rsLcDraws}{20}
\newcommand{\rsLcBonLo}{0.17}
\newcommand{\rsLcBonHi}{1.20}
\newcommand{\rsLcNguLo}{0.17}
\newcommand{\rsLcNguHi}{1.39}
\newcommand{\rsLcIndBetterN}{7}
\newcommand{\rsLcIndWorseN}{4}
\newcommand{\rsLcSeqLRBetterN}{16}
\newcommand{\rsLcSeqRLBetterN}{14}
\newcommand{\rsLcBonWBetterN}{13}
\newcommand{\rsLcIndWorseNmin}{30}
\newcommand{\rsLcIndWorseMax}{0.06}
\newcommand{\rsLcIndBetterNmax}{40}
\newcommand{\rsPctLcBonLo}{17}
\newcommand{\rsPctLcBonHi}{55}
\newcommand{\rsPctLcNguLo}{17}
\newcommand{\rsPctLcNguHi}{63}
\newcommand{\rsPctLcLitLo}{17}
\newcommand{\rsPctLcLitHi}{63}
\newcommand{\rsPctSmallIndLo}{76}
\newcommand{\rsPctSmallIndHi}{79}
\newcommand{\rsPctLcIndWorseMax}{10}
\newcommand{\rsLcSmallDualGtwoU}{1.18}
\newcommand{\rsLcSmallIndGtwoU}{5.49}
\newcommand{\rsLcSmallDualGtwoL}{1.61}
\newcommand{\rsLcSmallIndGtwoL}{6.86}
\newcommand{\rsLcBonWGtwoFullMax}{0.01}
\newcommand{\rsLcBonWSmallLo}{0.18}
\newcommand{\rsLcBonWSmallHi}{0.84}
\newcommand{\rsLcBonWGmaxLLo}{0.12}
\newcommand{\rsLcBonWGmaxLHi}{0.24}
\newcommand{\rsLcMlBetterN}{8}
\newcommand{\rsLcMlSmallLo}{0.07}
\newcommand{\rsLcMlSmallHi}{0.86}
\newcommand{\rsBonN}{8}
\newcommand{\rsBonLo}{0.13}
\newcommand{\rsBonHi}{0.75}
\newcommand{\rsNguN}{8}
\newcommand{\rsNguSigN}{7}
\newcommand{\rsNguHi}{1.28}
\newcommand{\rsNguNsGtwoUL}{+0.06 [$-$0.08, +0.18]}
\newcommand{\rsNguGMaxDiff}{0.15}
\newcommand{\rsNguGSigN}{1}
\newcommand{\rsJointN}{16}
\newcommand{\rsJointSigN}{15}
\newcommand{\rsJointHi}{1.30}
\newcommand{\rsExtLo}{9.4}
\newcommand{\rsExtHi}{114.8}
\newcommand{\rsNguJShiftMax}{110.6}
\newcommand{\rsNguFwdGtwoU}{+2.3}
\newcommand{\rsNguFwdGtwoL}{$-$3.5}
\newcommand{\rsNguFwdGmaxU}{+5.5}
\newcommand{\rsNguFwdGmaxL}{+9.8}
\newcommand{\rsFwdAgree}{0.03}
\newcommand{\rsBaseRollMaxG}{3.1}
\newcommand{\rsAblNoWaistBetterN}{7}
\newcommand{\rsAblNoWaistHi}{0.73}
\newcommand{\rsAblNoBranchBetterN}{3}
\newcommand{\rsAblNoBranchHi}{0.17}
\newcommand{\rsAblNoWaistLo}{0.12}
\newcommand{\rsAblNoTiltGmaxLo}{0.08}
\newcommand{\rsAblNoTiltGmaxHi}{0.50}
\newcommand{\rsAblMlGtwoUL}{$-$0.04 [$-$0.08, $-$0.01]}
\newcommand{\rsGainCompleteSigN}{8}
\newcommand{\rsGainCompleteLo}{0.13}
\newcommand{\rsGainCompleteHi}{0.75}
\newcommand{\rsGainProdLo}{0.16}
\newcommand{\rsGainProdHi}{2.96}
\newcommand{\rsSchurGtwoLo}{0.45}
\newcommand{\rsSchurGtwoHi}{0.71}
\newcommand{\rsSchurGtwoMed}{0.60}
\newcommand{\rsSchurGtwoBelow}{36}
\newcommand{\rsSchurGtwoN}{40}
\newcommand{\rsWaistIdGtwo}{20}
\newcommand{\rsCorrDualGtwoLo}{0.55}
\newcommand{\rsCorrDualGtwoHi}{0.57}
\newcommand{\rsCorrIndGtwoLo}{0.55}
\newcommand{\rsCorrIndGtwoHi}{0.69}
\newcommand{\rsSchurGmaxLo}{0.44}
\newcommand{\rsSchurGmaxHi}{0.79}
\newcommand{\rsSchurGmaxMed}{0.70}
\newcommand{\rsSchurGmaxBelow}{4}
\newcommand{\rsSchurGmaxN}{8}
\newcommand{\rsWaistIdGmax}{4}
\newcommand{\rsCorrDualGmaxLo}{0.78}
\newcommand{\rsCorrDualGmaxHi}{0.79}
\newcommand{\rsCorrIndGmaxLo}{0.87}
\newcommand{\rsCorrIndGmaxHi}{0.93}
\newcommand{\rsCompGtwoUL}{$-$0.05 [$-$0.18, +0.09]}
\newcommand{\rsCompSigN}{7}
\newcommand{\rsCompLoadedLo}{0.48}
\newcommand{\rsCompLoadedHi}{1.13}
\newcommand{\rsCompUnloadedLo}{0.27}
\newcommand{\rsCompUnloadedHi}{0.55}
\newcommand{\rsDeflGtwoU}{2.25/2.75}
\newcommand{\rsWaistDeflGtwoU}{1.42/1.44}
\newcommand{\rsDeflGtwoL}{5.11/5.39}
\newcommand{\rsWaistDeflGtwoL}{2.19/2.25}
\newcommand{\rsDeflGmaxU}{2.40/3.07}
\newcommand{\rsWaistDeflGmaxU}{2.13/2.27}
\newcommand{\rsDeflGmaxL}{7.00/8.14}
\newcommand{\rsWaistDeflGmaxL}{4.80/5.05}
\newcommand{\rsCwGmaxU}{0.064\pm0.004}
\newcommand{\rsCtGmaxU}{0.156\pm0.011}
\newcommand{\rsCwGmaxL}{0.092\pm0.004}
\newcommand{\rsCtGmaxL}{0.139\pm0.009}
\newcommand{\rsCwGmaxRatio}{43}
\newcommand{\rsCwJbGtwoU}{0.052\pm0.016}
\newcommand{\rsCwJbGtwoL}{0.050\pm0.018}
\newcommand{\rsCompConsistN}{14}
\newcommand{\rsCompInconsist}{G2max right joint 1, G2max right joint 3}
\newcommand{\rsCWristGtwoL}{3.50, 8.92, 2.76}
\newcommand{\rsMassCorrMin}{0.997}
\newcommand{\rsMassGainLo}{0.83}
\newcommand{\rsMassGainHi}{1.68}
\newcommand{\rsOutPose}{51}
\newcommand{\rsOutZRemoved}{5.37}
\newcommand{\rsOutPoseErr}{1.23}
\newcommand{\rsOutMedian}{0.44}
\newcommand{\rsOutMaxZ}{4.00}
\newcommand{\rsOutSensMax}{0.02}
\newcommand{\rsOutMaxZG}{4.42}
\newcommand{\rsNValLo}{14}
\newcommand{\rsNValHi}{25}
\newcommand{\rsDropGmaxU}{9}
\newcommand{\rsDropGmaxL}{12}
\newcommand{\rsTrainDualLo}{0.29}
\newcommand{\rsTrainDualHi}{0.75}
\newcommand{\MdhRightArmGTwo}{$\alpha_{5}$\,=\,$\pi$/2}
\newcommand{\MdhRightArmGTwoMax}{$\alpha_{1}$\,=\,$\pi$}
\newcommand{\TabTop}{\specialrule{0.6pt}{0pt}{1.2pt}\specialrule{0.6pt}{0pt}{\belowrulesep}}
\newcommand{\TabBottom}{\specialrule{0.6pt}{\aboverulesep}{1.2pt}\specialrule{0.6pt}{0pt}{0pt}}

\begin{document}

\title{Whole-Body Elasto-Geometrical Calibration of Wheeled Humanoid Robots Using a Shared-Waist Dual-Chain Model}

\author{Pengxin Zha\,\orcidlink{0009-0000-5103-8621}, Yi Lu$^{*}$\,\orcidlink{0009-0001-6883-9174}, Haojun Ma\,\orcidlink{0000-0002-2369-037X}, Jun Zhu, Minghao Qiu\,\orcidlink{0009-0008-9529-174X}, Xiaoyan Wang\,\orcidlink{0000-0001-6448-1470}, Wei Shan\,\orcidlink{0009-0004-1919-6311}, Maoqing Yao\,\orcidlink{0009-0007-7846-738X}
\thanks{This work is being submitted to the IEEE for possible publication. Copyright may be transferred without notice, after which this version may no longer be accessible.}
\thanks{The authors are with AgiBot, China. $^{*}$Corresponding author: Yi Lu (e-mail: luyi@zhiyuan-robot.com).}
}

\markboth{}{Zha \MakeLowercase{\textit{et al.}}: Whole-Body Elasto-Geometrical Calibration of Wheeled Humanoid Robots}

\maketitle

\begin{abstract}
Wheeled humanoid robots carry both arms on one actuated waist, usually without torque sensors, so the accuracy of both end-effectors depends on a single waist model. After a first stage that estimates the tracker and target extrinsics, we identify the waist and both arms in one problem from the target positions on both flanges, measured by an optical tracker. The kinematic model is complete, with waist-base, parallel-axis, and branch parameters; arm compliances are driven by measured torques and waist compliances by gravity torques of the robot's mass model; and Gaussian priors with evidence-maximizing scales stabilize the estimate. Independent and sequential single-chain calibration are restrictions of this problem, and its waist information is the sum of the single-chain Schur complements. On two robots with different waists, each without and with a payload, calibration reduces the mean errors on held-out poses from \rsNomLo--\rsNomHi~mm to \rsDualGCLo--\rsDualGCHi~mm. For every training-set size from \rsLcNmin{} poses to all, the method has \rsPctLcLitLo--\rsPctLcLitHi\% lower mean errors than adaptations of two published whole-body calibration methods, mainly through the waist compliance. Independent calibration yields two waist models; using either for both arms raises the other arm's error to up to \rsIndOneWaistHi~mm.
\end{abstract}


\section{Introduction}\label{sec:intro}
\IEEEPARstart{H}{umanoid} robots are moving from research laboratories into industrial work~\cite{Bogue2025}. Wheeled humanoids replace the legs with a mobile base and carry both arms on an actuated waist. In dual-arm tasks, both end-effectors must be accurate in one common base frame, and both depend on the same waist: an error in the waist model moves both hands at once, while geometric deviations along each chain accumulate at the end-effectors~\cite{Bonnet2023, Nguyen2025}. Harmonic drives add compliance, so joint deflections under load add to the geometric errors~\cite{Bonnet2023, Kamali2019}. The waist joints carry the upper body and both arms, but on wheeled humanoids they usually have no torque sensors.

Industrial manipulators are calibrated with external instruments such as laser trackers~\cite{Lou2022}; the identifiable parameters~\cite{Khalil1991} of a linearized error model of the serial chain are then estimated from the measurements~\cite{Zhang2025}. Elasto-geometrical calibration adds joint compliances driven by the joint loads~\cite{Kamali2019, Deng2023, Zhang2025}; without joint torque sensors, the loads follow from a gravity model of the links, which can be identified from the same calibration poses~\cite{Deng2023} or together with the kinematic parameters~\cite{Zhang2025, Yuan2023}, and the pose of a joint axis also varies with the posture of the axes that follow it~\cite{Cui2026}. Systems of two manipulators are calibrated with a holistic error model of the link parameters and of the base, camera, and marker transforms~\cite{Chen2025}, or with one arm as the measuring instrument of the other~\cite{Xiong2021}; on an industrial dual-arm robot, calibrating the cameras first and then the kinematics was more effective than calibrating all parameters simultaneously~\cite{Stepanova2022}. The two arms of a humanoid, however, share the waist joints, which raises the question of how the shared parameters are identified.

Early humanoid calibration relied on onboard sensors: head-mounted stereo cameras for the upper body~\cite{Birbach2015}, self-touch with artificial skin to close a kinematic chain~\cite{Roncone2014}, and link orientation sensors with both feet flat on the floor for the joint offsets~\cite{Yamane2011}; Tanguy et~al.~\cite{Tanguy2018} calibrated the geometric parameters of the whole kinematic tree online. In simulation on iCub, calibrating a chain together with other chains that share joints with it improved the optimization results and the observability~\cite{Stepanova2019}. Tenhumberg and B\"auml~\cite{Tenhumberg2021} modeled the upper body of Agile Justin with joint and transversal elasticities under its own weight, and identified its Denavit--Hartenberg parameters and elasticities by least squares with priors from the end-effector positions of both arms, measured by an external tracking system, which reduced the mean position error from 21 to 3.1~mm.

Two recent works calibrate the whole body of a humanoid and evaluate it by cross-validation. Bonnet et~al.~\cite{Bonnet2023} calibrated the bipedal TALOS with motion capture and its joint torque sensors: the head--trunk and both arms were identified in one least-squares problem that shares the trunk parameters, elasticities were driven by the measured torques, and identifiable parameters were selected by a QR decomposition on simulated configurations; the cross-validated marker error was 2.2~mm RMS. Nguyen et~al.~\cite{Nguyen2025} calibrated TALOS without external instruments, from contacts of the gripper with a single plane: two foot-to-hand chains, whose four identifiable torso parameters are common, are identified geometrically with a parameterization in the Unified Robot Description Format (URDF), and the cross-validated plane-contact error dropped by a factor of 2.3 relative to the manufacturer's model. Neither work reports how much the shared parameters gain over identifying the chains independently or one after the other, and neither addresses a wheeled robot whose waist carries no torque sensors.

This article studies these questions on wheeled humanoids calibrated in production with an optical tracker. Its contributions are as follows.
\begin{itemize}
\item A shared-waist dual-chain elasto-geometrical calibration for wheeled humanoids with a general waist (any number of revolute or prismatic joints): two stages separate the tracker and target extrinsics from the body model; waist-base, parallel-axis~\cite{Hayati1985}, and branch parameters complete the modified Denavit--Hartenberg (MDH) model; waist compliances about and perpendicular to the joint axes are driven by the gravity torques of the mass model of the robot description; and Gaussian priors, whose scales maximize the evidence of the training data, stabilize the estimate. The identification runs in the production calibration software of the robots.
\item An identifiability analysis in which independent and sequential single-chain calibration are row and column restrictions of the same system, and the dual-chain waist information is exactly the sum of the two single-chain Schur complements.
\item Experiments on two robots with different waists, each without load and with a payload, that compare the dual-chain formulation with independent and sequential calibration and with adaptations of~\cite{Bonnet2023, Nguyen2025} on held-out poses, for training sets from \rsLcNmin{} poses to all, and that quantify the contribution of each part of the model.
\end{itemize}

\section{Whole-Body Kinematic and Compliance Model}\label{sec:problem}

\subsection{Kinematic Structure and Measurement}\label{subsec:frames}
\begin{figure*}[!t]
\centering
\includegraphics{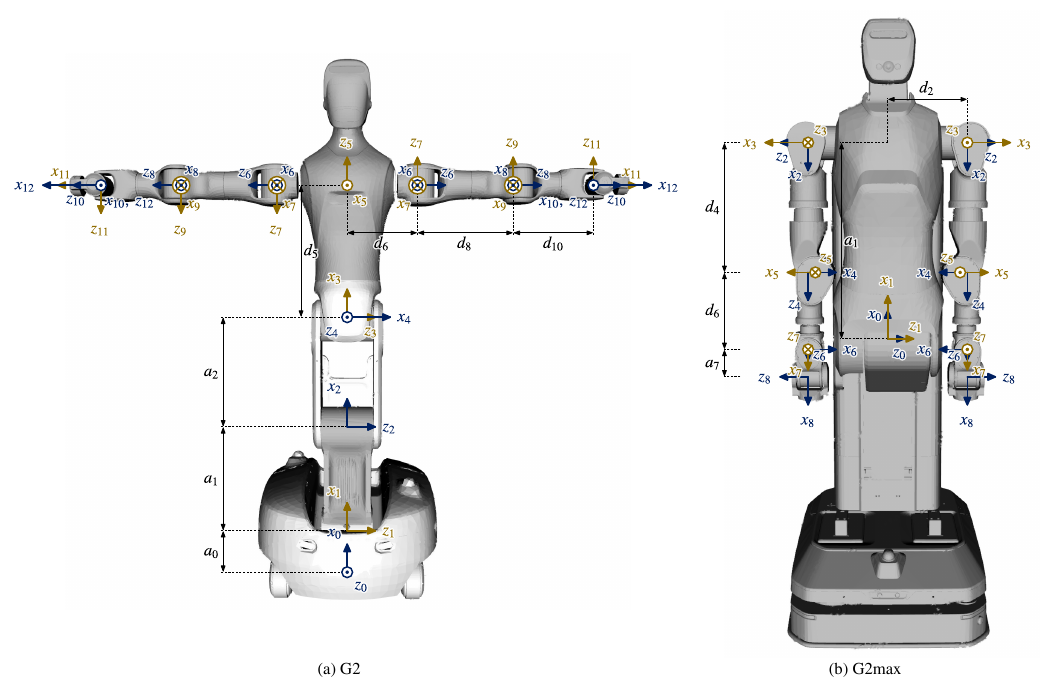}
\caption{MDH frames of the two robots at zero joint positions (front view), computed from the nominal model of the robot description. Arrows are the axes $x_i$ and $z_i$ in the drawing plane; a circle with a dot (a cross) is an axis pointing towards (away from) the viewer; colors alternate between consecutive frames. The symbols mark the link lengths $a_{i-1}$ and offsets $d_i$ visible in this view. (a) G2: joints 1--5 form the shared waist (frames 3 and 4 coincide), joints 6--12 belong to each arm. (b) G2max: the revolute pitch joint forms the waist (frames 0 and 1 coincide); joints 2--8 belong to each arm.}
\label{fig:system}
\end{figure*}

A wheeled humanoid consists of a mobile base, a waist with $n_w$ revolute or prismatic joints, and two arms with $n_a$ joints each. The waist is shared by the two chains, so each chain from the base to a flange has $n=n_w+n_a$ joints; arm joint $j$ is joint $n_w+j$ of the chain. The AGIBOT G2 (Fig.~\ref{fig:system}) has a waist of five revolute joints ($n_w=5$); the waist of the G2max is modeled by one revolute pitch joint ($n_w=1$). Both robots have two 7-DOF arms ($n_a=7$).

We define the tracker frame $\{s\}$, the base frame $\{0\}$ of the waist chain, and the target frames $\{b_l\}$ and $\{b_r\}$ on the left and right flanges. The MDH convention~\cite{Khalil1986} describes the transform between adjacent link frames,
\begin{equation}
\mathbf{T}_{i-1,i} = \mathbf{R}_x(\alpha_{i-1})\,\mathbf{T}_x(a_{i-1})\,\mathbf{R}_z(\theta_i)\,\mathbf{T}_z(d_i),
\label{eq:mdh}
\end{equation}
where $(\alpha_{i-1},a_{i-1},\theta_i,d_i)$ are the twist angle, link length, joint angle, and link offset, and the joint position is added to $\theta_i$ for a revolute joint and to $d_i$ for a prismatic joint. The target positions in $\{s\}$ follow from the two forward-kinematic chains
\begin{equation}
\begin{aligned}
\boldsymbol{y}_{b,k} = {}&[\mathbf{I}_3\ \mathbf{0}]\,\mathbf{T}_{s,0}\,\mathbf{T}_{\mathrm{wb}}\Bigl(\prod_{i=1}^{n_w}\mathbf{T}_{i-1,i}\Bigr)\mathbf{T}_{\mathrm{br},k}\\
&\cdot\Bigl(\prod_{i=n_w+1}^{n}\mathbf{T}_{i-1,i}^{k}\Bigr)\,\mathbf{T}_{n,b_k}\,\boldsymbol{t},\qquad k\in\{l,r\},
\end{aligned}
\label{eq:fk}
\end{equation}
where $[\mathbf{I}_3\ \mathbf{0}]$ extracts the position from a homogeneous coordinate and $\boldsymbol{t}=[0,0,0,1]^{\mathrm{T}}$. The waist factors are the same for both chains, and $\mathbf{T}_{s,0}\in\mathrm{SE}(3)$ is the tracker-to-base transform. The flange-to-target transform $\mathbf{T}_{n,b_k}=\mathbf{T}_{n,f_k}\,\mathbf{T}(\boldsymbol{p}_k)$ combines the constant flange transform of the robot description with the translation by the target offset $\boldsymbol{p}_k\in\mathbb{R}^3$. Only $\mathbf{T}_{s,0}$ and $\boldsymbol{p}_l,\boldsymbol{p}_r$ belong to the measurement setup; they form the extrinsic vector $\boldsymbol{\varepsilon}\in\mathbb{R}^{12}$ of Stage~I (Section~\ref{subsec:two_stage}).

Three additions, all zero in the nominal model, let the model represent any deviation of the joint axes from their nominal poses, i.e., make it complete in the sense of~\cite{Everett1987, Zhuang1992}. First, relative to $\{0\}$, the first waist axis is a line with four degrees of freedom, and the MDH row of the first waist joint covers only two of them: a tilt about and an offset along $x_0$, which is vertical on both robots. The waist-base transform $\mathbf{T}_{\mathrm{wb}}=\mathbf{T}_{u}(\delta_0)\,\mathbf{R}_{u}(\varphi_0)$ covers the other two, with the waist-base parameters $\boldsymbol{x}_{\mathrm{wb}}=[\delta_0,\ \varphi_0]^{\mathrm{T}}$: the offset along and the rotation about the horizontal direction $u$ normal to the axis, the forward direction of both robots. Second, if the axis of joint $i$ is nominally parallel to that of joint $i-1$, \eqref{eq:mdh} cannot describe a small rotation of axis $i$ about $y_{i-1}$; for such a joint, $\mathbf{R}_y(\beta_{i-1})$ precedes~\eqref{eq:mdh}~\cite{Hayati1985}, as in~\cite{Bonnet2023}. This concerns waist joints 2 and 3 of G2 and the first joint of each G2max arm, whose axis is parallel to the waist pitch axis. Third, both arms start at the frame of the last waist joint, so the shared factors cannot represent an offset or a rotation of one arm relative to the other along or about this axis. As for a link with several successors in~\cite{Khalil1986}, the first arm frame of chain $k$ is attached through the branch transform $\mathbf{T}_{\mathrm{br},k}=\mathbf{R}_z(\gamma_k)\,\mathbf{T}_z(b_k)$ with the branch parameters $(\gamma_k,b_k)$; one of the two pairs is redundant with $(\theta_{n_w},d_{n_w})$. Because $\mathbf{T}_{\mathrm{wb}}$ directly follows $\mathbf{T}_{s,0}$, an error of the Stage-I estimate of $\mathbf{T}_{s,0}$ changes only the six parameters of the first waist frame and does not propagate into the other body parameters; conversely, $\boldsymbol{x}_{\mathrm{wb}}$ cannot separate an offset of the waist mounting from such an error.

\subsection{Joint and Waist Compliance}\label{subsec:compliance}
With each harmonic-drive arm joint modeled as a linear torsional spring, the link-side joint positions of arm $k$ are
\begin{equation}
\tilde{\boldsymbol{q}}_{a,k} = \boldsymbol{q}_{a,k} - \mathrm{diag}(\boldsymbol{c}_k)\,\boldsymbol{\tau}_k,
\label{eq:compliance_joint}
\end{equation}
where $\boldsymbol{q}_{a,k},\boldsymbol{\tau}_k\in\mathbb{R}^{n_a}$ are the measured joint positions and the torques of the joint torque sensors in the static calibration poses, and $\boldsymbol{c}_k$ are the compliance coefficients.

The waist joints carry the upper body, both arms, and the payloads and deflect under this weight, like the elastic upper body of~\cite{Tenhumberg2021}, but they have no torque sensors. As for robots calibrated without joint torque sensors~\cite{Deng2023, Zhang2025, Yuan2023}, the loads are computed from a mass model; we take it from the robot description, with the payloads as point masses at the flanges, and do not identify it. For waist joint $j$ with origin $\boldsymbol{o}_j$ and axis $\boldsymbol{z}_j$ at the measured joint positions, the gravity moment of all distal links, with masses $m_i$ and centers of mass $\boldsymbol{r}_i$, is $\boldsymbol{\mu}_j = \sum_{i}(\boldsymbol{r}_i-\boldsymbol{o}_j)\times m_i\,\boldsymbol{g}$. Its component along the axis, $\tau_{w,j}=\boldsymbol{z}_j^{\mathrm{T}}\boldsymbol{\mu}_j$, twists the joint as in~\eqref{eq:compliance_joint}, with the waist compliance $c_{w,j}$: $\tilde{q}_{w,j}=q_{w,j}-c_{w,j}\tau_{w,j}$. Its component normal to the axis, $\boldsymbol{\mu}_{\perp,j}=\boldsymbol{\mu}_j-\boldsymbol{z}_j\tau_{w,j}$, bends the joint and its support, which we model as a small rotation $c_{t,j}\boldsymbol{\mu}_{\perp,j}$ of the distal body about $\boldsymbol{o}_j$, with the tilt compliance $c_{t,j}$. Elasticities transverse to the joint axes were also modeled in~\cite{Tenhumberg2021}, and posture-dependent deviations of the joint axes in~\cite{Cui2026}. Both compliances are shared by the two chains, and the tilts move the target of chain $k$ by
\begin{equation}
\Delta\boldsymbol{y}_{t,k}=\mathbf{R}_{s,0}\sum_{j=1}^{n_w} c_{t,j}\,\boldsymbol{\mu}_{\perp,j}\times(\boldsymbol{p}_{b,k}-\boldsymbol{o}_j),
\label{eq:tilt}
\end{equation}
where $\boldsymbol{p}_{b,k}$ is the target position in $\{0\}$ and $\mathbf{R}_{s,0}$ the rotation of $\mathbf{T}_{s,0}$. The moments are evaluated with the nominal kinematics. The same mass model gives the torques of the arm joints, which are measured: for arm joints 1--4, the modeled and measured torques have correlation coefficients of at least \rsMassCorrMin{} on all datasets, and the measured torques are \rsMassGainLo--\rsMassGainHi{} times the modeled ones; a constant scale error of the waist torques only scales the identified waist compliances.

The predicted target position $\hat{\boldsymbol{y}}_k$ evaluates~\eqref{eq:fk} at the link-side positions $\tilde{\boldsymbol{q}}_{w}$ and $\tilde{\boldsymbol{q}}_{a,k}$ and adds~\eqref{eq:tilt}. For small deflections, its sensitivity to the arm compliances is $\mathbf{J}_{c,k}=-\mathbf{J}_{v,k}\mathrm{diag}(\boldsymbol{\tau}_k)$, where $\mathbf{J}_{v,k}$ is the translational Jacobian of the target with respect to the arm joints; the sensitivities $\mathbf{J}_{w,k}$ to the waist compliances $\boldsymbol{c}_w,\boldsymbol{c}_t\in\mathbb{R}^{n_w}$ follow in the same way from the waist joints and~\eqref{eq:tilt}.

\section{Dual-Chain Identification}\label{sec:method}
\subsection{Two-Stage Procedure}\label{subsec:two_stage}
\begin{figure*}[!t]
\centering
\includegraphics{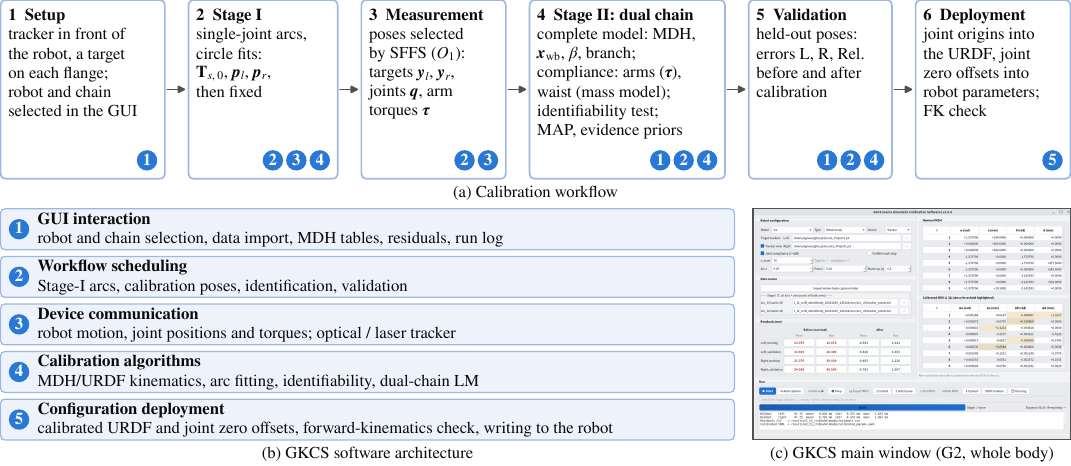}
\caption{Overall calibration workflow and its implementation in GKCS. (a) Steps of the procedure (Sections~\ref{sec:method} and~\ref{sec:exp}); circled numbers: GKCS layers of (b) that run the step. (b) Layered architecture of GKCS with the functions of each layer. (c) GKCS main window after the whole-body calibration of G2 without load: calibration data, errors before and after calibration on the training and validation poses, nominal MDH parameters and identified deviations, and the identification log.}
\label{fig:workflow}
\end{figure*}

Fig.~\ref{fig:workflow}(a) shows the complete procedure. In Stage~I, single joints move along arcs, and circles fitted to the target trajectories give the joint axes (circle point analysis~\cite{Stone1987}), as in the frame identification of industrial robots~\cite{Ferrarini2024}. $\mathbf{T}_{s,0}$ follows from the axes of the waist joints 3--5 on G2 and, with the waist locked at zero, from the axes of the left arm joints 1 and 2 on G2max; $\boldsymbol{p}_l$ and $\boldsymbol{p}_r$ follow from arcs of the arm joints 6 and 7. In Stage~II, $\boldsymbol{\varepsilon}$ is held fixed, and the body parameters and compliances are identified from the static calibration poses. We do not estimate the extrinsics jointly with the body parameters: the base transform and the target offsets would trade off against the proximal and distal body parameters, so the converged body parameters would depend on where the tracker and the targets are placed. The waist-base parameters are identifiable with $\boldsymbol{\varepsilon}$ fixed, because no MDH parameter represents their two directions.

\subsection{Parameter Vector and Dual-Chain Error Model}\label{subsec:params}
The body parameters identified in Stage~II are
\begin{equation}
\boldsymbol{e} = \bigl[\boldsymbol{x}_s^{\mathrm{T}}\ \ \boldsymbol{x}_{a,l}^{\mathrm{T}}\ \ \boldsymbol{x}_{a,r}^{\mathrm{T}}\ \ \boldsymbol{c}_l^{\mathrm{T}}\ \ \boldsymbol{c}_r^{\mathrm{T}}\ \ \boldsymbol{c}_w^{\mathrm{T}}\ \ \boldsymbol{c}_t^{\mathrm{T}}\bigr]^{\mathrm{T}}\in\mathbb{R}^{n_e},
\label{eq:x_full}
\end{equation}
where $\boldsymbol{x}_s\in\mathbb{R}^{n_s}$ collects the shared waist parameters (MDH, $\beta_{i-1}$, and $\boldsymbol{x}_{\mathrm{wb}}$), $\boldsymbol{x}_{a,l},\boldsymbol{x}_{a,r}\in\mathbb{R}^{n_x}$ the arm parameters (MDH, $\beta_{i-1}$, and branch), and $n_e=n_s+2n_x+2n_a+2n_w$ ($n_s=24$, $n_x=30$, and $n_e=108$ for G2; $n_s=6$, $n_x=31$, and $n_e=84$ for G2max). Selection matrices route the shared block to both chains: $\boldsymbol{x}_k=[\boldsymbol{x}_s^{\mathrm{T}}\ \boldsymbol{x}_{a,k}^{\mathrm{T}}]^{\mathrm{T}}=\mathbf{S}_k\boldsymbol{x}$ with $\boldsymbol{x}=[\boldsymbol{x}_s^{\mathrm{T}}\ \boldsymbol{x}_{a,l}^{\mathrm{T}}\ \boldsymbol{x}_{a,r}^{\mathrm{T}}]^{\mathrm{T}}$. Differentiating~\eqref{eq:fk} with respect to $\boldsymbol{x}_k$ gives the geometric Jacobian $\mathbf{J}_{x,k}\in\mathbb{R}^{3\times(n_s+n_x)}$, and one pose with both targets measured yields the linearized dual-chain model
\begin{equation}
\begin{bmatrix}
\Delta \boldsymbol{y}_l \\
\Delta \boldsymbol{y}_r
\end{bmatrix}=\underbrace{\begin{bmatrix}
\mathbf{J}_{x,l}\mathbf{S}_l & \mathbf{J}_{c,l} & \mathbf{0} & \mathbf{J}_{w,l} \\
\mathbf{J}_{x,r}\mathbf{S}_r & \mathbf{0} & \mathbf{J}_{c,r} & \mathbf{J}_{w,r}
\end{bmatrix}}_{\mathbf{H}\in\mathbb{R}^{6\times n_e}}\Delta\boldsymbol{e},
\label{eq:H}
\end{equation}
where $\Delta\boldsymbol{y}_k=\boldsymbol{y}_k^{\mathrm{meas}}-\hat{\boldsymbol{y}}_k$. The shared columns of both chains, including those of the waist compliances, map to the same slot of $\boldsymbol{e}$, while $\boldsymbol{c}_l$ and $\boldsymbol{c}_r$ enter only their own chain. Stacking~\eqref{eq:H} over $N$ poses gives $\Delta\tilde{\boldsymbol{y}}=\tilde{\mathbf{H}}\,\Delta\boldsymbol{e}$ with $\tilde{\mathbf{H}}\in\mathbb{R}^{6N\times n_e}$. The geometry-only variant (G) drops all compliance columns; the elasto-geometrical variant (GC) keeps them.
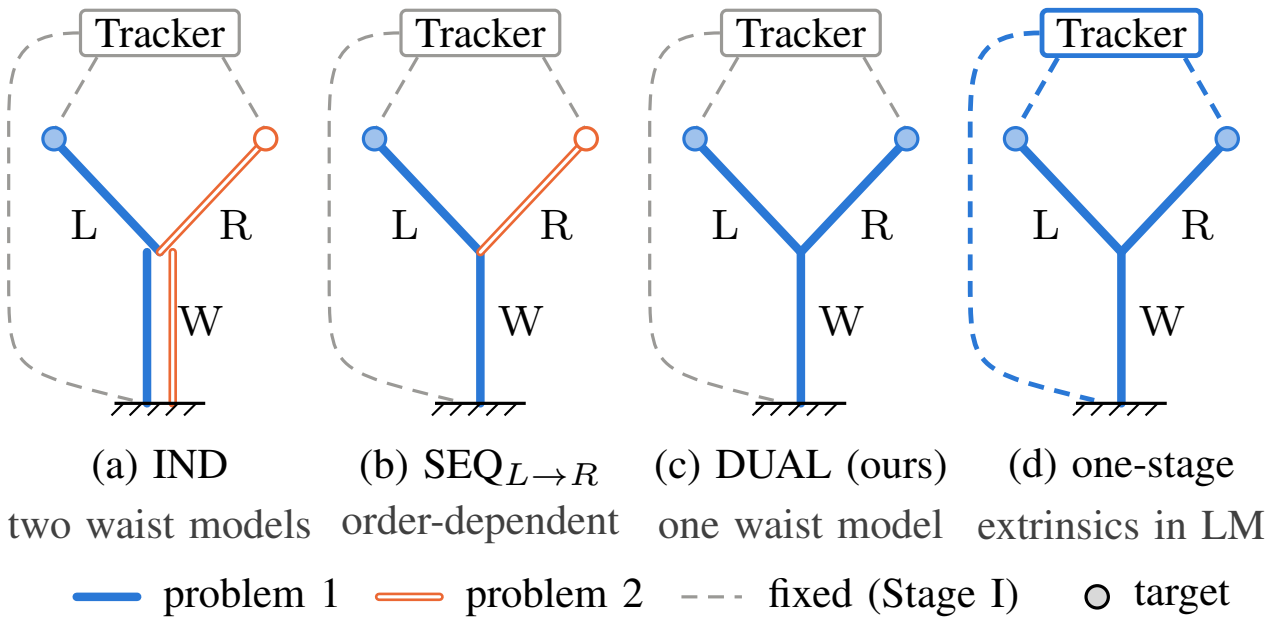
\begin{figure}[!t]
\centering
\definecolor{cA}{HTML}{2A78D6}%
\definecolor{cB}{HTML}{EB6834}%
\definecolor{cF}{HTML}{9A9996}%
\begin{tikzpicture}[x=1cm, y=1cm, font=\footnotesize,
    link/.style={line width=1.6pt, line cap=round},
    meas/.style={densely dashed, line width=0.6pt},
    tgt/.style={circle, draw, line width=0.6pt, inner sep=0pt, minimum size=4.2pt},
    trk/.style={draw, line width=0.6pt, rounded corners=1pt, minimum width=1.05cm, minimum height=0.30cm, inner sep=1pt},
    p1/.style={draw=cA},
    p2/.style={draw=cB, double=white, line width=0.5pt, double distance=0.6pt},
    p1 tgt/.style={draw=cA, fill=cA!45},
    p2 tgt/.style={draw=cB, fill=white},
    fx/.style={draw=cF},
    ex1/.style={draw=cA, line width=0.9pt}]
  \newcommand{\ychain}[8]{%
    \begin{scope}[xshift=#1cm]
      \node[trk, #6, text=black] (T) at (0, 2.45) {Tracker};
      \draw[meas, #6] (T.south west) ++(0.12,0) -- (-0.70, 1.80);
      \draw[meas, #6] (T.south east) ++(-0.12,0) -- (0.70, 1.80);
      \draw[meas, #6] (T.west) .. controls (-0.95, 2.45) and (-1.0, 2.25) .. (-1.0, 1.85)
            -- (-1.0, 0.85) .. controls (-1.0, 0.22) .. (-0.18, 0.02);
      \ifx\relax#3\relax
        \draw[link, #2] (0, 0) -- (0, 1.0);
      \else
        \draw[link, #2] (-0.085, 0) -- (-0.085, 1.0);
        \draw[link, #3] (0.085, 0) -- (0.085, 1.0);
      \fi
      \draw[link, #4] (0, 1.0) -- (-0.66, 1.70);
      \draw[link, #5] (0, 1.0) -- (0.66, 1.70);
      \node[tgt, #4 tgt] at (-0.70, 1.75) {};
      \node[tgt, #5 tgt] at (0.70, 1.75) {};
      \draw[line width=0.6pt] (-0.30, 0) -- (0.30, 0);
      \foreach \k in {-0.24, -0.12, 0, 0.12, 0.24} \draw[line width=0.4pt] (\k, 0) -- ++(-0.08, -0.08);
      \node at (0.27, 0.55) {$\mathrm{W}$};
      \node at (-0.50, 1.18) {$\mathrm{L}$};
      \node at (0.50, 1.18) {$\mathrm{R}$};
      \node[align=center, text width=2.1cm] at (0, -0.42) {#7};
      \node[align=center, text width=2.1cm, text=black!75] at (0, -0.80) {#8};
    \end{scope}}
  \ychain{0}{p1}{p2}{p1}{p2}{fx}{(a) IND}{two waist models}
  \ychain{2.12}{p1}{}{p1}{p2}{fx}{(b) SEQ$_{L\to R}$}{order-dependent}
  \ychain{4.24}{p1}{}{p1}{p1}{fx}{(c) DUAL (ours)}{one waist model}
  \ychain{6.36}{p1}{}{p1}{p1}{ex1}{(d) one-stage}{extrinsics in LM}
  \begin{scope}[yshift=-1.28cm]
    \draw[link, p1] (-0.55, 0) -- (-0.15, 0);  \node[anchor=west] at (-0.10, 0) {problem 1};
    \draw[link, p2] (1.45, 0) -- (1.85, 0);  \node[anchor=west] at (1.90, 0) {problem 2};
    \draw[meas, fx] (3.45, 0) -- (3.85, 0);  \node[anchor=west] at (3.90, 0) {fixed (Stage~I)};
    \node[tgt, draw=black, fill=black!15] at (6.20, 0) {};  \node[anchor=west] at (6.32, 0) {target};
  \end{scope}
\end{tikzpicture}
\caption{Compared formulations. Colors mark the least-squares problem in which a part of the model is estimated and the target data it uses; gray dashed lines are the extrinsics, fixed from Stage~I. The waist (W) includes the waist-base parameters and the waist compliances. (a) IND identifies each chain alone and returns two waist models. (b) SEQ$_{L\to R}$ identifies the waist with the left arm and then the right arm with the waist fixed. (c) DUAL identifies the waist and both arms in one problem; the adaptations BON and NGU have the same structure with other body models. (d) One-stage estimation of the extrinsics with the body parameters, as in the original procedures of~\cite{Bonnet2023, Nguyen2025}.}
\label{fig:methods}
\end{figure}

\subsection{Single-Chain Formulations as Restrictions}\label{subsec:baselines}
Independent and sequential calibration select rows and columns of $\tilde{\mathbf{H}}$ (Fig.~\ref{fig:methods}). Independent calibration (IND) solves two problems, one per target, each with the columns of the waist ($\boldsymbol{x}_s$ and the waist compliances) and of one arm ($\boldsymbol{x}_{a,k}$ and $\boldsymbol{c}_k$). It returns two estimates of the one waist; we evaluate each arm with its own chain, which favors IND, because a robot can use only one waist model. Sequential calibration SEQ$_{L\to R}$ first solves the left chain as in IND and then the right arm alone, with the rows of the right target and the columns of $(\boldsymbol{x}_{a,r},\boldsymbol{c}_r)$; SEQ$_{R\to L}$ is the mirror order. The dual-chain formulation (DUAL) uses all rows and columns. All formulations share the identifiability test, the solver, and the stopping rule of Section~\ref{sec:lm}.

\subsection{Waist Identifiability}\label{subsec:excitation}
Order the columns of $\tilde{\mathbf{H}}$ as waist ($\boldsymbol{x}_s$ and, in variant GC, the waist compliances), left-arm, and right-arm blocks ($\boldsymbol{x}_{a,k}$ and, in variant GC, $\boldsymbol{c}_k$). With the rows of the left and right targets,
\begin{equation}
\tilde{\mathbf{H}} = \begin{bmatrix} \mathbf{A}_l & \mathbf{B}_l & \mathbf{0} \\ \mathbf{A}_r & \mathbf{0} & \mathbf{B}_r \end{bmatrix}.
\label{eq:blocks}
\end{equation}
With independent measurement noise of variance $\sigma^2$, the marginal covariance of the waist estimate is $\sigma^2\mathbf{F}_w^{-1}$, where $\mathbf{F}_w$ is the Schur complement of the arm blocks in $\tilde{\mathbf{H}}^{\mathrm{T}}\tilde{\mathbf{H}}$. Because the two arm blocks share neither rows nor columns, the elimination decouples:
\begin{equation}
\mathbf{F}_w^{\mathrm{dual}} = \mathbf{F}_l + \mathbf{F}_r,\quad
\mathbf{F}_k = \mathbf{A}_k^{\mathrm{T}}\bigl(\mathbf{I}-\mathbf{P}_{k}\bigr)\mathbf{A}_k,
\label{eq:schur}
\end{equation}
where $\mathbf{P}_{k}=\mathbf{B}_k(\mathbf{B}_k^{\mathrm{T}}\mathbf{B}_k)^{-1}\mathbf{B}_k^{\mathrm{T}}$ projects onto the identifiable columns of arm block $k$, and $\mathbf{F}_k$ is the waist information of the single chain $k$ in IND. Eq.~\eqref{eq:schur} is the additivity of the information of independent measurements, applied to the waist after the arm parameters are eliminated; it gives in closed form the gain of combining chains that share joints, which~\cite{Stepanova2019} observed in simulation. Hence the dual-chain waist covariance $\sigma^2(\mathbf{F}_l+\mathbf{F}_r)^{-1}$ is bounded by that of either single chain in the Loewner order. If $\mathbf{F}_l$ and $\mathbf{F}_r$ are proportional, the ratio of the dual-chain standard deviation of a parameter to that of the better chain is at least $1/\sqrt{2}$, with equality for equal information, as for doubled data; ratios below $1/\sqrt{2}$ therefore indicate that the two arms contribute different waist directions.

\subsection{Identifiability Test and Solver}\label{sec:lm}
Not all columns of $\tilde{\mathbf{H}}$ are identifiable~\cite{Khalil1991}. At the nominal parameters, a Gauss--Jordan elimination processes the columns in the order waist (MDH and $\beta_{i-1}$), branch, arms (MDH and $\beta_{i-1}$), base, and compliances (arms, waist), and keeps a column only if its pivot exceeds $10^{-4}\max_{ij}|\tilde{H}_{ij}|$; the other columns are frozen at their initial values. This removes the branch parameters of the right arm as redundant with the shared waist row. In the implementation, $\mathbf{T}_{\mathrm{wb}}$ is realized by refining $\mathbf{T}_{s,0}$: its six columns follow all body columns, and the elimination retains exactly two of them in every fit, which together with the body columns span the same column space as the body columns and $\boldsymbol{x}_{\mathrm{wb}}$; the other four directions of $\mathbf{T}_{s,0}$ and the target offsets stay at their Stage-I values. The columns are evaluated by central differences.

The free parameters have Gaussian priors on their deviations $\boldsymbol{u}$ from the initial values, $\boldsymbol{u}_g\sim\mathcal{N}(\mathbf{0},\lambda_g^{-1}\mathbf{I})$ for each group $g$: lengths, twist and $\beta$ angles, joint offsets, compliances of arm joints 1--4, compliances of arm joints 5--7, waist compliances, and tilt compliances; $\boldsymbol{x}_{\mathrm{wb}}$ has a flat prior. Priors were also used in~\cite{Tenhumberg2021}. Here, the precisions $\lambda_g$ and the noise variance $\sigma^2$ are not tuned by hand but maximize the evidence, i.e., the marginal likelihood of the training data~\cite{MacKay1992}. With the posterior covariance $\mathbf{S}=(\tilde{\mathbf{H}}^{\mathrm{T}}\tilde{\mathbf{H}}/\sigma^2+\boldsymbol{\Lambda})^{-1}$ and $\boldsymbol{\Lambda}=\mathrm{diag}(\lambda_{g(k)})$ at the current estimate, the updates are
\begin{equation}
\gamma_k = 1-\lambda_{g(k)}S_{kk},\quad
\lambda_g=\frac{\sum_{k\in g}\gamma_k}{\|\boldsymbol{u}_g\|^2},\quad
\sigma^2=\frac{\|\Delta\tilde{\boldsymbol{y}}\|^2}{6N-\sum_k\gamma_k},
\label{eq:evidence}
\end{equation}
where $\gamma_k\in[0,1]$ measures how far parameter $k$ is determined by the data. They alternate with the maximum a posteriori (MAP) estimate, which minimizes $\|\Delta\tilde{\boldsymbol{y}}\|^2/\sigma^2+\boldsymbol{u}^{\mathrm{T}}\boldsymbol{\Lambda}\boldsymbol{u}$, starting from the least-squares estimate with $\boldsymbol{\Lambda}=\mathbf{0}$, until the precisions change by less than 0.1\% or for at most 30 rounds. A group that the data do not determine, such as the wrist compliances without load, receives a large precision and stays at zero; with many poses, the priors become weak and the estimate approaches least squares.

Each estimate is computed with Levenberg--Marquardt (LM) steps with Marquardt's diagonal scaling~\cite{Marquardt1963}, on the residuals and the regressor augmented by the prior terms; the iteration stops when the relative cost reduction falls below $10^{-9}$, the step norm below $10^{-10}$, or after 150 iterations. The MDH parameters start at their nominal values, all other parameters at zero. Predictions use the exact forward kinematics with~\eqref{eq:compliance_joint} and~\eqref{eq:tilt}.

\section{Experiments}\label{sec:exp}
\subsection{Platforms, Measurement, and Data}\label{subsec:platform}
\begin{figure}[!t]
\centering
\includegraphics{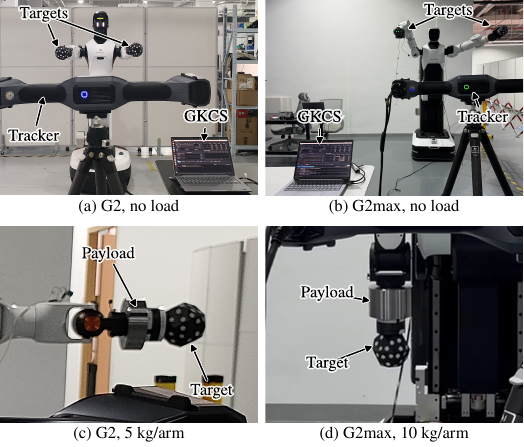}
\caption{Experimental setup. (a) G2 and (b) G2max without payload: the optical tracker in front of the robot measures the targets on both flanges, and GKCS runs on the laptop. (c) 5~kg payload and target on a G2 flange; (d) 10~kg payload and target on a G2max flange.}
\label{fig:setup}
\end{figure}

The experiments use one G2 and one G2max. Each arm joint has a joint torque sensor. An optical 3-D tracker (FreeScan Trak Nova, SHINING~3D) measures a marker target, a ball covered with retro-reflective dots, on each flange (Fig.~\ref{fig:setup}); in an acceptance test according to ISO~10360-8:2013, its length measurement error was within the maximum permissible error of $\pm0.062$~mm. The payload of 5~kg per arm (G2) or 10~kg per arm (G2max) moves the target farther from the flange than without load. All measurements are run by the calibration software GKCS (Genie Kinematic Calibration Software, Section~\ref{subsec:deploy}).

The calibration poses were selected per robot by sequential forward floating selection~\cite{Pudil1994, Deng2023}, which maximizes the $O_1$ observability index~\cite{Joubair2013} of the dual-chain Jacobian over random candidates that satisfy the joint limits and keep both targets visible. Of the 100 selected poses, those that caused interference on the robot were removed, leaving 95 for G2 and 81 for G2max; with the 10~kg payloads, 9 of the 81 G2max poses caused interference and were not executed, leaving 72. Each robot was measured without load and with the payload (Table~\ref{tab:datasets}); each dataset has its own Stage~I, split, and calibration, because the tracker station and the target position differ between the sessions. The poses are split in the order of the pose list, as the production software does: on G2, poses 1--70 for training and 71--95 for validation; on G2max, poses 1--60 and 61--81.
\begin{table}[!t]
\caption{Datasets. Waist: Joints of the Waist Model (R: Revolute); Poses: Executed Calibration Poses; Used: After Cleaning; $|\tau_1|$: Mean Absolute Torque of Arm Joint~1.}
\label{tab:datasets}
\centering
\footnotesize
\setlength{\tabcolsep}{3.2pt}

\begin{tabular}{lcccccc}
\TabTop
\textbf{Dataset} & \textbf{Waist} & \textbf{Payload} & \textbf{Poses} & \textbf{Used} & \textbf{Train/val} & $|\boldsymbol{\tau}_{1}|$ \textbf{L/R} \\
 & & (kg/arm) & & & & (N$\cdot$m) \\
\midrule
G2, no load & 5R & 0 & 95 & 95 & 70/25 & 15.3/14.8 \\
G2, loaded & 5R & 5 & 95 & 95 & 70/25 & 36.7/35.9 \\
G2max, no load & 1R & 0 & 81 & 71 & 54/17 & 33.1/47.2 \\
G2max, loaded & 1R & 10 & 72 & 60 & 46/14 & 90.0/87.9 \\
\TabBottom
\end{tabular}
\end{table}

All methods use the same cleaned data, obtained with two rules. Missing data: a pose is removed if a target position, a joint position, or a torque of either arm is missing, which happens when the tracker loses a target (\rsDropGmaxU{} and \rsDropGmaxL{} poses on G2max, none on G2). Gross errors: DUAL-GC is fitted to all remaining poses of a dataset, and a pose is removed if the larger error $r$ of its two targets exceeds the median by more than five robust standard deviations, $(r-\mathrm{median})/(1.4826\,\mathrm{MAD})>5$, where MAD is the median absolute deviation, with up to three passes. This rule removes one pose, pose~\rsOutPose{} of the unloaded G2max (score \rsOutZRemoved), whose larger target error is \rsOutPoseErr~mm against a median of \rsOutMedian~mm; the largest score of the remaining poses is \rsOutMaxZ. Screening with DUAL-G (largest score \rsOutMaxZG), or no screening, keeps this pose and changes the mean validation errors of DUAL-GC by at most \rsOutSensMax~mm. The G2max sessions continue with the poses of a separate test after the calibration poses; these are not used.

\subsection{Compared Methods and Metrics}\label{subsec:compared}
IND, SEQ$_{L\to R}$, SEQ$_{R\to L}$, and DUAL use the same model and estimator, in both variants G and GC. The components of the model beyond the complete kinematics, the waist and tilt compliances and the priors, were chosen by cross-validation on the training poses of each dataset; the validation poses are used only for the evaluation. Two whole-body methods from the literature are adapted to the same data and poses. For a controlled comparison, both use the two flange targets as observations, instead of motion-capture markers on most links~\cite{Bonnet2023} or plane contacts~\cite{Nguyen2025}, use the Stage-I extrinsics, and can represent the complete pose of the first waist axis, so that only the body model and its identification differ; their posture-selection algorithms would require new measurements and are not reproduced.
\begin{itemize}
\item BON-GC, after~\cite{Bonnet2023}: MDH parameters with $\beta$ for consecutive parallel axes, arm compliances driven by the measured torques, and identifiable parameters selected by a QR decomposition of the regressor on 1000 random configurations within the joint limits. The six columns of $\mathbf{T}_{s,0}$ enter the decomposition after the joint parameters, as in~\cite{Bonnet2023}, which keeps two of them, as for DUAL; the target offsets stay fixed. As~\cite{Bonnet2023} does not specify the solver, we use the LM solver of Section~\ref{sec:lm} without priors. The trunk of TALOS has joint torque sensors, which drive its compliances in~\cite{Bonnet2023}; our waist has none, so BON-GC has no waist compliance. BON-GC$^{w}$ adds the waist and tilt compliances of Section~\ref{subsec:compliance}, which separates their effect from that of the other parts of the proposed method.
\item NGU-G, after~\cite{Nguyen2025}: URDF-style parameters (six per joint origin), shared waist parameters, geometry only, a pivoted QR decomposition, and the same solver. The origin of the first waist joint already contains the waist base, so all extrinsics stay fixed. At the nominal parameters, the decomposition loses the directions of exactly parallel axes (on G2max, arm joint~1 and the waist pitch), so we repeat it at the current estimate until the selected set no longer changes. The plane-contact constraints, central to~\cite{Nguyen2025}, are replaced by the tracker measurements, so this adaptation tests the parameterization, not the plane-contact measurement.
\end{itemize}
The metrics are the Euclidean position errors of the left and right targets (L, R) and the dual-arm relative error $\|(\hat{\boldsymbol{y}}_l-\hat{\boldsymbol{y}}_r)-(\boldsymbol{y}_l^{\mathrm{meas}}-\boldsymbol{y}_r^{\mathrm{meas}})\|$ (Rel.). Differences between methods are evaluated per validation pose, with 95\% confidence intervals (CIs) from a paired bootstrap ($10^4$ resamples). For smaller training sets, every method is also fitted to \rsLcDraws{} random subsets of $N$ training poses, $N=20,25,30,40,50,60$ where available, the same subsets for all methods, and evaluated on the validation poses; the differences to DUAL-GC, pooled over both arms, are tested with a bootstrap that resamples both the subsets and the validation poses. With the full training sets, this gives \rsLcN{} cases of dataset and training-set size.

\subsection{Single- and Dual-Chain Formulations}\label{subsec:results}
Table~\ref{tab:errors} lists the mean errors on the validation poses. Calibration reduces them from \rsNomLo--\rsNomHi~mm (nominal model) to \rsDualGCGtwoLo--\rsDualGCGtwoHi~mm on G2 and \rsDualGCGmaxLo--\rsDualGCGmaxHi~mm on G2max (DUAL-GC; \rsTrainDualLo--\rsTrainDualHi~mm on the training poses). The identifiability test keeps \rsFreeGtwoG{} parameters (G) and \rsFreeGtwoGC{} (GC) on G2 and \rsFreeGmaxG{} and \rsFreeGmaxGC{} on G2max, and the evidence assigns DUAL-GC \rsGammaLo--\rsGammaHi{} effective parameters $\sum_k\gamma_k$. Tables~\ref{tab:mdh_g2} and~\ref{tab:mdh_g2max} list the nominal parameters and those identified by DUAL-GC on the datasets without load.
\begin{table*}[!t]
\caption{Mean Position Errors on the Validation Poses (mm). L, R: Left and Right Target; Rel.: Dual-Arm Relative Error. Bold: Lowest Calibrated Value per Column. $^{+}$/$^{-}$: Larger/Smaller than DUAL of the Same Variant (BON: DUAL-GC; NGU: DUAL-G), Paired 95\% CI Excluding Zero. IND, SEQ, and DUAL Use the Proposed Model; BON-GC$^{w}$: BON-GC with the Waist Compliances of the Proposed Model. All Methods Use the Stage-I Extrinsics and a Complete Waist Base.}
\label{tab:errors}
\centering
\footnotesize
\setlength{\tabcolsep}{3.6pt}

\begin{tabular}{lcccccccccccc}
\TabTop
 & \multicolumn{3}{c}{\textbf{G2, no load}} & \multicolumn{3}{c}{\textbf{G2, 5\,kg/arm}} & \multicolumn{3}{c}{\textbf{G2max, no load}} & \multicolumn{3}{c}{\textbf{G2max, 10\,kg/arm}} \\
\cmidrule(lr){2-4} \cmidrule(lr){5-7} \cmidrule(lr){8-10} \cmidrule(lr){11-13}
\textbf{Method} & \textbf{L} & \textbf{R} & \textbf{Rel.} & \textbf{L} & \textbf{R} & \textbf{Rel.} & \textbf{L} & \textbf{R} & \textbf{Rel.} & \textbf{L} & \textbf{R} & \textbf{Rel.} \\
\midrule
Nominal & 13.92 & 24.60 & 21.50 & 17.21 & 31.14 & 22.60 & 22.16 & 11.31 & 25.01 & 25.11 & 16.35 & 27.99 \\
IND-G & 0.84 & 1.10 & 0.89 & 1.50 & 1.62 & 1.82 & 1.11 & 0.78 & 1.21 & 2.04 & 1.54 & 2.20 \\
IND-GC & 0.82 & 0.82 & 0.80 & \textbf{0.99}$^{-}$ & 1.05$^{+}$ & \textbf{1.19} & \textbf{0.55} & \textbf{0.45} & \textbf{0.60} & \textbf{0.65} & \textbf{0.59} & \textbf{0.82} \\
SEQ$_{L\to R}$-G & 0.84 & 1.20$^{+}$ & 0.98 & 1.50 & 1.91$^{+}$ & 2.08$^{+}$ & 1.11 & 0.86 & 1.34 & 2.04 & 2.55$^{+}$ & 2.55$^{+}$ \\
SEQ$_{L\to R}$-GC & 0.82 & 0.88$^{+}$ & 0.80 & \textbf{0.99}$^{-}$ & 1.21$^{+}$ & 1.39$^{+}$ & \textbf{0.55} & 0.67$^{+}$ & 0.83$^{+}$ & \textbf{0.65} & 1.52$^{+}$ & 1.56$^{+}$ \\
SEQ$_{R\to L}$-G & 0.97 & 1.10 & 1.00 & 1.86$^{+}$ & 1.62 & 2.01$^{+}$ & 1.20$^{+}$ & 0.78 & 1.27 & 2.07 & 1.54 & 2.29 \\
SEQ$_{R\to L}$-GC & \textbf{0.80} & 0.82 & 0.77 & 1.29$^{+}$ & 1.05$^{+}$ & 1.26 & 0.76$^{+}$ & \textbf{0.45} & 0.78$^{+}$ & 1.09$^{+}$ & \textbf{0.59} & 1.09$^{+}$ \\
DUAL-G & 0.88 & 1.07 & 0.93 & 1.59 & 1.55 & 1.84 & 1.16 & 0.78 & 1.24 & 1.84 & 1.67 & 2.10 \\
DUAL-GC & 0.83 & \textbf{0.79} & \textbf{0.73} & 1.11 & 0.97 & 1.20 & 0.61 & 0.50 & 0.69 & 0.72 & 0.64 & 0.94 \\
\addlinespace[1pt]
BON-GC~\cite{Bonnet2023} & 0.99$^{+}$ & 0.97$^{+}$ & 0.97$^{+}$ & 1.44$^{+}$ & 1.38$^{+}$ & 1.58$^{+}$ & 1.02$^{+}$ & 0.63$^{+}$ & 0.88 & 1.46$^{+}$ & 1.39$^{+}$ & 1.43$^{+}$ \\
BON-GC$^{w}$~\cite{Bonnet2023} & \textbf{0.80} & 0.80 & 0.81$^{+}$ & 1.12 & \textbf{0.93} & 1.24 & 0.64 & 0.51 & 0.78 & 0.87$^{+}$ & 0.73 & 1.15 \\
NGU-G~\cite{Nguyen2025} & 0.88 & 1.08 & 0.92 & 1.57 & 1.57 & 1.83 & 1.18 & 0.84$^{+}$ & 1.25 & 2.00 & 1.70 & 2.15 \\
\TabBottom
\end{tabular}
\end{table*}
\begin{table*}[!t]
\caption{MDH Parameters of G2: Nominal (Robot Description, Zero Joint Positions) and Identified by DUAL-GC on the Dataset Without Load. $c_i$: Compliance About the Joint Axis, $c_{t,i}$: Tilt Compliance of a Waist Joint, Both in $10^{-4}$~rad/(N$\cdot$m); $\beta_{i-1}$ Only for Axes Parallel to the Previous One (--: Not in the Model); \textbackslash: Not Identifiable and Kept at the Nominal Value; $(\gamma_l,b_l)$: Branch Parameters of the Left Arm (Those of the Right Arm Are Redundant). The Right Arm Shares the Waist of the Left Chain; Its Nominal Parameters Differ Only in \MdhRightArmGTwo}
\label{tab:mdh_g2}
\centering
\footnotesize
\setlength{\tabcolsep}{2pt}

\begin{tabular*}{\textwidth}{@{\extracolsep{\fill}}cl*{12}{c}@{}}
\TabTop
 & & \multicolumn{5}{c}{\textbf{Waist}} & \multicolumn{7}{c}{\textbf{Arm}} \\
\cmidrule(lr){3-7} \cmidrule(l){8-14}
 & \textbf{Parameters} & \textbf{Joint 1} & \textbf{Joint 2} & \textbf{Joint 3} & \textbf{Joint 4} & \textbf{Joint 5} & \textbf{Joint 6} & \textbf{Joint 7} & \textbf{Joint 8} & \textbf{Joint 9} & \textbf{Joint 10} & \textbf{Joint 11} & \textbf{Joint 12} \\
\midrule
\multirow{5}{*}{\rotatebox[origin=c]{90}{\footnotesize Nominal}} & $a_{i-1}$ (mm) & 144 & 360 & 380 & 0 & 0 & 0 & 0 & 0 & 10.5 & 10.5 & 0 & 0 \\
 & $\alpha_{i-1}$ (rad) & $\pi$/2 & 0 & 0 & $-\pi$/2 & $-\pi$/2 & $-\pi$/2 & $\pi$/2 & $\pi$/2 & $\pi$/2 & $\pi$/2 & $\pi$/2 & $\pi$/2 \\
 & $d_i$ (mm) & 0 & 0 & 0 & 0 & 457.5 & 243.5 & 0 & 332 & 0 & 278 & 0 & 0 \\
 & $\theta_i$ (rad) & 0 & 0 & 0 & $-\pi$/2 & $-\pi$/2 & 0 & $-\pi$ & $-\pi$ & $-\pi$ & $-\pi$ & $\pi$/2 & 0 \\
 & $\beta_{i-1}$ (mrad) & -- & 0 & 0 & -- & -- & -- & -- & -- & -- & -- & -- & -- \\
\midrule
\multirow{7}{*}{\rotatebox[origin=c]{90}{\footnotesize Left chain}} & $a_{i-1}$ (mm) & 144.620 & 360.070 & 381.423 & $-$0.271 & $-$0.683 & 5.854 & 0.133 & $-$0.050 & 10.574 & 10.318 & 0.304 & $-$0.026 \\
 & $\alpha_{i-1}$ (rad) & 1.5706 & 0.0003 & $-$0.0008 & $-$1.5717 & $-$1.5699 & $-$1.5710 & 1.5693 & 1.5723 & 1.5717 & 1.5689 & 1.5718 & 1.5710 \\
 & $d_i$ (mm) & 1.323 & \textbackslash & \textbackslash & $-$0.523 & 457.789 & 243.866 & $-$0.078 & 332.165 & 0.383 & 278.411 & $-$0.100 & $-$0.628 \\
 & $\theta_i$ (rad) & $-$0.0085 & 0.0109 & $-$0.0048 & $-$1.5706 & $-$1.5785 & 0.0046 & $-$3.1405 & $-$3.1442 & $-$3.1391 & $-$3.1449 & 1.5754 & 0.0035 \\
 & $\beta_{i-1}$ (mrad) & -- & 0.11 & 1.90 & -- & -- & -- & -- & -- & -- & -- & -- & -- \\
 & $c_i$ & $-$0.005 & 0.052 & 0.014 & 0.005 & 0.015 & 0.41 & 0.68 & 1.59 & 1.42 & 0.00 & 0.00 & 0.00 \\
 & $c_{t,i}$ & 0.016 & 0.018 & 0.015 & 0.027 & 0.036 & -- & -- & -- & -- & -- & -- & -- \\
\midrule
\multirow{5}{*}{\rotatebox[origin=c]{90}{\footnotesize Right arm}} & $a_{i-1}$ (mm) & \multicolumn{5}{c}{} & 5.650 & 0.094 & 0.108 & 10.702 & 10.419 & 0.306 & $-$0.291 \\
 & $\alpha_{i-1}$ (rad) & \multicolumn{5}{c}{} & 1.5713 & 1.5724 & 1.5707 & 1.5701 & 1.5730 & 1.5716 & 1.5710 \\
 & $d_i$ (mm) & \multicolumn{5}{c}{shared} & 243.501 & 0.225 & 332.218 & $-$0.078 & 278.681 & 0.328 & $-$3.529 \\
 & $\theta_i$ (rad) & \multicolumn{5}{c}{} & $-$0.0354 & $-$3.1442 & $-$3.1297 & $-$3.1460 & $-$3.1432 & 1.5511 & $-$0.0360 \\
 & $c_i$ & \multicolumn{5}{c}{} & 0.87 & 0.69 & 1.97 & 1.68 & 0.00 & 0.00 & 0.00 \\
\midrule
\multicolumn{2}{@{}l}{$\boldsymbol{x}_{\mathrm{wb}}$ (mm, mrad)} & \multicolumn{12}{c}{(3.39, $-$0.34)} \\
\multicolumn{2}{@{}l}{$(\gamma_l, b_l)$ (mrad, mm)} & \multicolumn{12}{c}{($-$0.43, 0.38)} \\
\TabBottom
\end{tabular*}
\end{table*}
\begin{table*}[!t]
\caption{MDH Parameters of G2max: Nominal and Identified by DUAL-GC on the Dataset Without Load (Notation of Table~\ref{tab:mdh_g2}). The Nominal Right Arm Differs Only in \MdhRightArmGTwoMax}
\label{tab:mdh_g2max}
\centering
\footnotesize
\setlength{\tabcolsep}{2pt}

\begin{tabular*}{\textwidth}{@{\extracolsep{\fill}}cl*{8}{c}@{}}
\TabTop
 & & \multicolumn{1}{c}{\textbf{Waist}} & \multicolumn{7}{c}{\textbf{Arm}} \\
\cmidrule(lr){3-3} \cmidrule(l){4-10}
 & \textbf{Parameters} & \textbf{Joint 1} & \textbf{Joint 2} & \textbf{Joint 3} & \textbf{Joint 4} & \textbf{Joint 5} & \textbf{Joint 6} & \textbf{Joint 7} & \textbf{Joint 8} \\
\midrule
\multirow{5}{*}{\rotatebox[origin=c]{90}{\footnotesize Nominal}} & $a_{i-1}$ (mm) & 0 & 680 & 0 & 0 & 25 & 25 & 0 & 95 \\
 & $\alpha_{i-1}$ (rad) & 0 & 0 & $\pi$/2 & $\pi$/2 & $\pi$/2 & $\pi$/2 & $\pi$/2 & $-\pi$/2 \\
 & $d_i$ (mm) & 0 & 276 & 0 & 450 & 0 & 267 & 0 & 0 \\
 & $\theta_i$ (rad) & 0 & $-\pi$ & $\pi$/2 & $-\pi$ & $-\pi$ & $-\pi$ & $\pi$/2 & 0 \\
 & $\beta_{i-1}$ (mrad) & -- & 0 & -- & -- & -- & -- & -- & -- \\
\midrule
\multirow{7}{*}{\rotatebox[origin=c]{90}{\footnotesize Left chain}} & $a_{i-1}$ (mm) & $-$0.896 & 679.763 & 0.016 & $-$0.075 & 25.046 & 25.293 & 0.068 & 95.144 \\
 & $\alpha_{i-1}$ (rad) & $-$0.0003 & $-$0.0014 & 1.5709 & 1.5715 & 1.5709 & 1.5693 & 1.5706 & $-$1.5701 \\
 & $d_i$ (mm) & 0.288 & \textbackslash & 0.127 & 450.250 & 0.122 & 267.203 & 0.090 & $-$1.278 \\
 & $\theta_i$ (rad) & $-$0.0097 & $-$3.1347 & 1.5729 & $-$3.1918 & $-$3.1391 & $-$3.1548 & 1.5742 & $-$0.0029 \\
 & $\beta_{i-1}$ (mrad) & -- & $-$0.46 & -- & -- & -- & -- & -- & -- \\
 & $c_i$ & 0.064 & 0.25 & 0.18 & 0.26 & 0.17 & 0.94 & 0.29 & 0.12 \\
 & $c_{t,i}$ & 0.156 & -- & -- & -- & -- & -- & -- & -- \\
\midrule
\multirow{6}{*}{\rotatebox[origin=c]{90}{\footnotesize Right arm}} & $a_{i-1}$ (mm) & \multicolumn{1}{c}{} & 679.349 & $-$0.234 & $-$0.006 & 24.956 & 25.240 & $-$0.031 & 94.990 \\
 & $\alpha_{i-1}$ (rad) & \multicolumn{1}{c}{} & 3.1404 & 1.5708 & 1.5704 & 1.5706 & 1.5727 & 1.5703 & $-$1.5713 \\
 & $d_i$ (mm) & \multicolumn{1}{c}{shared} & \textbackslash & 0.069 & 450.284 & $-$0.130 & 267.162 & $-$0.157 & 0.549 \\
 & $\theta_i$ (rad) & \multicolumn{1}{c}{} & $-$3.1589 & 1.5633 & $-$3.1339 & $-$3.1323 & $-$3.1899 & 1.5645 & 0.0090 \\
 & $\beta_{i-1}$ (mrad) & \multicolumn{1}{c}{} & $-$0.22 & -- & -- & -- & -- & -- & -- \\
 & $c_i$ & \multicolumn{1}{c}{} & 0.19 & 0.23 & 0.14 & 0.19 & $-$0.17 & 1.14 & $-$0.42 \\
\midrule
\multicolumn{2}{@{}l}{$\boldsymbol{x}_{\mathrm{wb}}$ (mm, mrad)} & \multicolumn{8}{c}{(6.72, 0.09)} \\
\multicolumn{2}{@{}l}{$(\gamma_l, b_l)$ (mrad, mm)} & \multicolumn{8}{c}{(1.64, 0.17)} \\
\TabBottom
\end{tabular*}
\end{table*}

Sequential calibration loses accuracy on the arm identified second: this arm is less accurate than with DUAL in \rsSeqWorseN{} of the \rsSeqN{} combinations of dataset, variant, and order, significantly in \rsSeqSigN{} (\rsSeqGCSigN{} of \rsSeqGCN{} in variant GC), by up to \rsSeqHi~mm, and never significantly more accurate; the arm identified first equals IND by construction.

With all training poses, independent calibration reaches per-arm accuracies close to DUAL: of the \rsIndN{} per-arm differences (at most \rsIndMaxDiff~mm), \rsIndWorseN{} is significantly in favor of IND and \rsIndBetterN{} in favor of DUAL, both on the loaded G2 (left \rsIndGCLoadedL~mm, right \rsIndGCLoadedR~mm, IND-GC minus DUAL-GC), and none of the relative errors differs significantly. The two waist models of IND, however, disagree: over all poses, the waist end frames of the left and right estimates differ on average by \rsIndPosGtwoLo--\rsIndPosGtwoHi~mm and \rsIndRotGtwoLo--\rsIndRotGtwoHi~mrad on G2 and by \rsIndPosGmaxLo--\rsIndPosGmaxHi~mm and \rsIndRotGmaxLo--\rsIndRotGmaxHi~mrad on G2max (variant GC). If the waist model of one chain, including its waist compliances, is used for both arms, the mean error of the other arm rises from \rsIndOwnWaistLo--\rsIndOwnWaistHi~mm to \rsIndOneWaistLo--\rsIndOneWaistHi~mm; refitting the other arm on this waist is SEQ.

\begin{figure}[!t]
\centering
\includegraphics{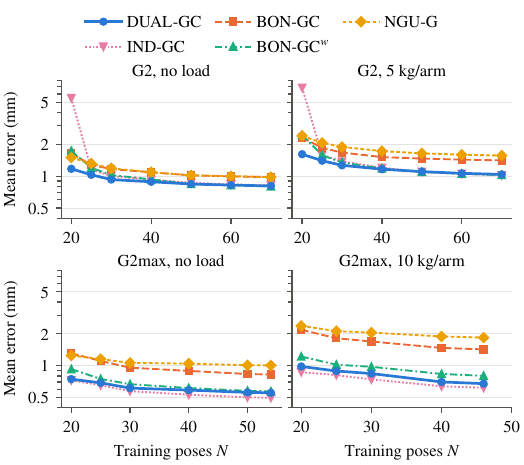}
\caption{Mean validation error of both targets against the number of training poses ($N$; \rsLcDraws{} random subsets per size, the full training set once), for DUAL-GC and IND-GC with the proposed model and for the adaptations of~\cite{Bonnet2023, Nguyen2025}; BON-GC$^{w}$ adds the waist compliances of the proposed model to BON-GC. Logarithmic scale.}
\label{fig:learning}
\end{figure}

With fewer training poses, DUAL is also more accurate per arm (Fig.~\ref{fig:learning}): with \rsLcNmin{} poses, the mean errors of IND-GC are \rsLcSmallIndGtwoU{} and \rsLcSmallIndGtwoL~mm on G2, against \rsLcSmallDualGtwoU{} and \rsLcSmallDualGtwoL~mm (\rsPctSmallIndLo--\rsPctSmallIndHi\% lower) for DUAL-GC, because a single chain must identify the five-joint waist from its own target alone. Over the \rsLcN{} cases, DUAL-GC is significantly more accurate than IND-GC in \rsLcIndBetterN{}, all on G2 with up to \rsLcIndBetterNmax{} poses, and less accurate in \rsLcIndWorseN{}, all on the unloaded G2max with \rsLcIndWorseNmin{} or more poses, where the errors of IND-GC are lower by at most \rsLcIndWorseMax~mm (\rsPctLcIndWorseMax\%). SEQ is never significantly more accurate than DUAL-GC and less accurate in \rsLcSeqLRBetterN{} (SEQ$_{L\to R}$) and \rsLcSeqRLBetterN{} (SEQ$_{R\to L}$) cases.

\subsection{Waist Identifiability}\label{subsec:res_ident}
\begin{figure}[!t]
\centering
\includegraphics{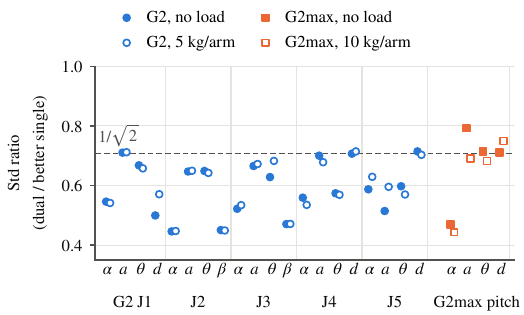}
\caption{Standard deviation of each identifiable waist parameter in DUAL relative to the better single chain, from~\eqref{eq:schur} at the DUAL-GC solution; the waist block also contains $\boldsymbol{x}_{\mathrm{wb}}$ and the waist compliances. The dashed line marks $1/\sqrt{2}$, the ratio for two chains with equal information.}
\label{fig:waist}
\end{figure}

The identifiability test keeps the same waist parameters in the single and dual chains: \rsWaistIdGtwo{} on G2 (18 of the 20 MDH parameters, as the parallel axes of joints 1--3 make two link offsets redundant, and $\beta_1$, $\beta_2$) and all \rsWaistIdGmax{} MDH parameters on G2max, together with both waist-base parameters and, in variant GC, all waist compliances in every fit. The dual chain lowers their uncertainty. Fig.~\ref{fig:waist} shows~\eqref{eq:schur} at the DUAL-GC solution, with $\boldsymbol{x}_{\mathrm{wb}}$ and the waist compliances in the waist block and the information of the data only, without the priors. Over both payloads, the dual-chain standard deviations of the waist parameters are \rsSchurGtwoLo--\rsSchurGtwoHi{} (median \rsSchurGtwoMed) times those of the better single chain on G2 and \rsSchurGmaxLo--\rsSchurGmaxHi{} (median \rsSchurGmaxMed) on G2max. On G2, \rsSchurGtwoBelow{} of \rsSchurGtwoN{} ratios lie below $1/\sqrt{2}$, so the two arms contribute complementary information; on G2max, \rsSchurGmaxBelow{} of \rsSchurGmaxN{} do. The largest correlation between a waist parameter and a shoulder parameter (arm joints 1--2) changes little: from \rsCorrIndGtwoLo--\rsCorrIndGtwoHi{} (IND) to \rsCorrDualGtwoLo--\rsCorrDualGtwoHi{} (DUAL) on G2 and from \rsCorrIndGmaxLo--\rsCorrIndGmaxHi{} to \rsCorrDualGmaxLo--\rsCorrDualGmaxHi{} on G2max.

\subsection{Whole-Body Methods from the Literature}\label{subsec:literature}
With the same Stage-I extrinsics and all training poses, DUAL-GC is more accurate than BON-GC on all \rsBonN{} arms, by \rsBonLo--\rsBonHi~mm (all significant, Table~\ref{tab:errors}), and than NGU-G on \rsNguSigN{} of \rsNguN{} arms, by up to \rsNguHi~mm; the remaining arm, the left one of the unloaded G2, differs by \rsNguNsGtwoUL~mm (NGU-G minus DUAL-GC). NGU-G reaches the errors of DUAL-G, whose geometric model class it shares (differences of at most \rsNguGMaxDiff~mm, \rsNguGSigN{} of 8 significant). The learning curves (Fig.~\ref{fig:learning}) show the same for every training-set size: in all \rsLcN{} cases, DUAL-GC is significantly more accurate than BON-GC, by \rsLcBonLo--\rsLcBonHi~mm (\rsPctLcBonLo--\rsPctLcBonHi\% of its mean error), and than NGU-G, by \rsLcNguLo--\rsLcNguHi~mm (\rsPctLcNguLo--\rsPctLcNguHi\%).

The difference comes mainly from the waist compliance. BON-GC$^{w}$, which adds it to BON-GC, reaches the errors of DUAL-GC on G2 with all training poses (within \rsLcBonWGtwoFullMax~mm, not significant), and it is never significantly more accurate than DUAL-GC. DUAL-GC is more accurate in \rsLcBonWBetterN{} of the \rsLcN{} cases: with \rsLcNmin{} poses by \rsLcBonWSmallLo--\rsLcBonWSmallHi~mm, where the priors stabilize the estimate, and on the loaded G2max at every size, by \rsLcBonWGmaxLLo--\rsLcBonWGmaxLHi~mm, where removing the branch parameters from DUAL-GC has a similar effect (Section~\ref{subsec:ablation}).

NGU-G and DUAL take up the same offset of the waist base: NGU-G moves the origin of its first waist joint forward by \rsNguFwdGtwoU, \rsNguFwdGtwoL, \rsNguFwdGmaxU, and \rsNguFwdGmaxL~mm (G2 and G2max, each without and with load), and the forward offsets $\delta_0$ of DUAL-G agree with these within \rsFwdAgree~mm, with rolls $\varphi_0$ of at most \rsBaseRollMaxG~mrad. They differ between the two sessions of each robot, so they contain the Stage-I error of the base transform and should not be read as properties of the robot.

With the extrinsics estimated together with the body parameters, as in~\cite{Bonnet2023, Nguyen2025}, BON-GC and NGU-G are less accurate than DUAL-GC on \rsJointSigN{} of the \rsJointN{} arms, by up to \rsJointHi~mm, and in all \rsLcN{} cases of the learning curves. The base transform and the first waist joint are then partly redundant, and their split depends on the initialization rather than on the robot: the tracker-to-base translations end \rsExtLo--\rsExtHi~mm away from the Stage-I estimates, and the origin of the first waist joint of NGU moves by up to \rsNguJShiftMax~mm. The two-stage procedure avoids this dependence of the body parameters on the measurement setup.

\subsection{Parts of the Model}\label{subsec:ablation}
Table~\ref{tab:ablation} removes one part of the proposed model at a time. The waist compliance has the largest effect: without it, the errors of DUAL-GC rise by \rsAblNoWaistLo--\rsAblNoWaistHi~mm, significantly on \rsAblNoWaistBetterN{} of 8 arms. The tilt compliance matters mainly on G2max (+\rsAblNoTiltGmaxLo{} to +\rsAblNoTiltGmaxHi~mm without it) and the branch parameters on \rsAblNoBranchBetterN{} arms, by up to \rsAblNoBranchHi~mm. With all training poses, the priors change the errors little: least squares is significantly more accurate on one arm, the left arm of the unloaded G2 (\rsAblMlGtwoUL~mm), and less accurate on none. With few poses, however, the priors are essential: with \rsLcNmin{} poses, least squares raises the errors by \rsLcMlSmallLo--\rsLcMlSmallHi~mm, and over the \rsLcN{} cases, the priors are significantly better in \rsLcMlBetterN{} and never worse. Against the models with arm compliances only and least squares, the proposed model lowers the errors by \rsGainCompleteLo--\rsGainCompleteHi~mm (complete kinematic model, significant on \rsGainCompleteSigN{} of 8 arms) and by \rsGainProdLo--\rsGainProdHi~mm (MDH model without waist base).
\begin{table}[!t]
\caption{Validation Errors (mm) of DUAL-GC with Parts of the Proposed Model Removed, All Training Poses. Compl.: Compliance; $^{+}$/$^{-}$: Larger/Smaller than the Proposed Model, Paired 95\% CI Excluding Zero. Last Three Rows: Arm Compliances Only and Least Squares, with Complete Kinematics, Without $\beta$ and Branch, and Also Without Waist Base.}
\label{tab:ablation}
\centering
\footnotesize
\setlength{\tabcolsep}{1.8pt}

\begin{tabular}{lcccccccc}
\TabTop
 & \multicolumn{2}{c}{\textbf{G2}} & \multicolumn{2}{c}{\textbf{G2 5\,kg}} & \multicolumn{2}{c}{\textbf{G2max}} & \multicolumn{2}{c}{\textbf{G2max 10\,kg}} \\
\cmidrule(lr){2-3} \cmidrule(lr){4-5} \cmidrule(lr){6-7} \cmidrule(lr){8-9}
\textbf{Model} & \textbf{L} & \textbf{R} & \textbf{L} & \textbf{R} & \textbf{L} & \textbf{R} & \textbf{L} & \textbf{R} \\
\midrule
Proposed & 0.83 & 0.79 & 1.11 & 0.97 & 0.61 & 0.50 & 0.72 & 0.64 \\
Least squares & 0.78$^{-}$ & 0.82 & 1.09 & 0.95 & 0.60 & 0.48 & 0.71 & 0.66 \\
No tilt compl. & 0.83 & 0.84$^{+}$ & 1.15 & 1.05$^{+}$ & 0.79$^{+}$ & 0.58 & 1.22$^{+}$ & 0.78$^{+}$ \\
No waist compl. & 0.96$^{+}$ & 0.96$^{+}$ & 1.43$^{+}$ & 1.34$^{+}$ & 0.87$^{+}$ & 0.62 & 1.40$^{+}$ & 1.37$^{+}$ \\
No branch & 0.85$^{+}$ & 0.78 & 1.15$^{+}$ & 0.96 & 0.60 & 0.57 & 0.89$^{+}$ & 0.69 \\
No $\beta$, no branch & 0.77 & 0.83 & 1.11 & 1.01 & 0.59 & 0.60$^{+}$ & 0.91$^{+}$ & 0.72 \\
\addlinespace[1pt]
Arm compl.\ only & 0.96$^{+}$ & 0.98$^{+}$ & 1.41$^{+}$ & 1.37$^{+}$ & 0.97$^{+}$ & 0.65$^{+}$ & 1.47$^{+}$ & 1.38$^{+}$ \\
\quad no $\beta$, branch & 0.91 & 1.00$^{+}$ & 1.37$^{+}$ & 1.37$^{+}$ & 1.02$^{+}$ & 0.64$^{+}$ & 1.47$^{+}$ & 1.25$^{+}$ \\
\quad no $\boldsymbol{x}_{\mathrm{wb}}$ & 0.98$^{+}$ & 1.00$^{+}$ & 1.49$^{+}$ & 1.64$^{+}$ & 2.18$^{+}$ & 1.73$^{+}$ & 3.10$^{+}$ & 3.60$^{+}$ \\
\TabBottom
\end{tabular}
\end{table}

\subsection{Joint and Waist Compliance}\label{subsec:res_compliance}
Comparing the variants GC and G of DUAL pose by pose, the compliances reduce the errors on \rsCompSigN{} of 8 arms significantly, by \rsCompLoadedLo--\rsCompLoadedHi~mm under load and by \rsCompUnloadedLo--\rsCompUnloadedHi~mm on three arms without load; the left arm of the unloaded G2 changes by \rsCompGtwoUL~mm. The identified compliances move the targets of the validation poses by \rsDeflGtwoU~mm (left/right) on the unloaded and \rsDeflGtwoL~mm on the loaded G2 and by \rsDeflGmaxU{} and \rsDeflGmaxL~mm on G2max; the waist compliances alone account for \rsWaistDeflGtwoU, \rsWaistDeflGtwoL, \rsWaistDeflGmaxU, and \rsWaistDeflGmaxL~mm. The gains in error are smaller than these displacements because the geometric parameters of DUAL-G absorb their pose-independent part.

On G2max, the waist pitch has the tilt compliance $c_t=\rsCtGmaxU$ without and $\rsCtGmaxL$ with load, which agree, and the axial compliance $c_w=\rsCwGmaxU$ and $\rsCwGmaxL$ (in $10^{-4}$~rad/(N$\cdot$m), $\pm1\sigma$ of the posterior), which is \rsCwGmaxRatio\% larger with load; a linear spring loaded by a point-mass payload does not explain this difference. On G2, only the compliance of waist joint~2 about its axis exceeds two standard deviations, with the same value in both sessions ($\rsCwJbGtwoU$ and $\rsCwJbGtwoL$); the other waist compliances, including all tilt compliances, are within two standard deviations of zero. The compliances of arm joints 1--4 agree between the two payloads within two standard deviations of their difference in \rsCompConsistN{} of 16 cases (exceptions: \rsCompInconsist). Without load, the evidence keeps the wrist compliances of G2 at zero ($\gamma_k\approx0$); with load, they are identified (left arm, joints 5--7: \rsCWristGtwoL). Whether compliances identified at one payload predict the errors at another is not tested here.

\subsection{Calibration Software and Production Deployment}\label{subsec:deploy}
GKCS has a layered architecture of GUI interaction, workflow scheduling, device communication, calibration algorithms, and configuration deployment, which together run all steps of the procedure (Fig.~\ref{fig:workflow}). GKCS writes the calibrated joint origins into the robot description (URDF) and the corrected joint zero offsets into the robot parameters, so that the controller and the planners use the calibrated model without further conversion. A whole-body calibration takes about 20~min, against more than 1~h before. GKCS has been deployed in production at the factories of the manufacturer, where it has calibrated more than one hundred G2 and several tens of G2max robots and has considerably improved the absolute positioning accuracy of their whole bodies.

\section{Discussion}\label{sec:discussion}
On these datasets, the proposed method has \rsPctLcLitLo--\rsPctLcLitHi\% lower mean errors than the adaptations of~\cite{Bonnet2023, Nguyen2025} for every training-set size, mostly owing to the waist compliance: with it, the adaptation of~\cite{Bonnet2023} reaches DUAL-GC on G2 with all training poses. The priors add accuracy with few poses, and the branch parameters on the loaded G2max. Independent calibration is more accurate per arm on the unloaded G2max with \rsLcIndWorseNmin{} or more poses (by at most \rsLcIndWorseMax~mm or \rsPctLcIndWorseMax\% in the learning curves), but it yields two waist models that cannot be used together.

The evaluation has limits. Each dataset is split once, into training poses and \rsNValLo--\rsNValHi{} held-out poses of the same session; the poses were selected for the dual-chain Jacobian, the parts of the model were chosen with the training poses of the same sessions, and the CIs are not corrected for multiple comparisons. Repeated sessions of the same robot under the same condition and task-level tests such as dual-arm positioning are not evaluated. The waist loads come from the CAD mass model with the payloads as point masses, and the tilt compliance lumps the bending of the joints and of their supports. The separation of the measurement setup is not complete either: residual Stage-I errors of the base transform are taken up by the first waist frame, and those of the target offsets can only be taken up by the last arm frames. Prismatic waist joints are covered by the model but not tested experimentally. The adaptations use our measurements and poses, not the sensors and posture-selection methods of~\cite{Bonnet2023, Nguyen2025}.

\section{Conclusion}\label{sec:conclusion}
This article presented a shared-waist dual-chain elasto-geometrical calibration for wheeled humanoid robots with a general waist, with a complete kinematic model, waist compliances driven by the gravity torques of the mass model, and priors whose scales maximize the evidence of the training data. On two robots with two payload conditions each, calibration reduced the mean validation errors from \rsNomLo--\rsNomHi~mm to \rsDualGCLo--\rsDualGCHi~mm. For every training-set size from \rsLcNmin{} poses to all, the mean errors of the method were \rsPctLcLitLo--\rsPctLcLitHi\% lower than those of adaptations of~\cite{Bonnet2023, Nguyen2025}, mainly because of the waist compliance. Compared with single-chain calibration, the dual-chain formulation avoided the accuracy loss of the arm identified second, replaced two inconsistent waist models by one, lowered the waist uncertainty, and, with \rsLcNmin{} training poses on G2, had \rsPctSmallIndLo--\rsPctSmallIndHi\% lower errors than independent calibration. The identification runs in production in GKCS. Future work will add waist arcs to Stage~I of G2max and test the identified compliances across payloads.

\end{document}